\documentclass[11pt]{article}
\usepackage{geometry}                % See geometry.pdf to learn the layout options. There are lots.
\usepackage{graphicx,url}
\usepackage{amssymb}
\usepackage{natbib}
\usepackage{amsmath}
\usepackage{mathtools}
\usepackage{epstopdf}
\usepackage{algorithm}
\usepackage{algpseudocode}
\usepackage{caption}
\usepackage{bm}% bold math
\usepackage{bbm}
\usepackage[auth-sc]{authblk}
\newcommand{\be}{\begin{eqnarray}}
\newcommand{\ee}{\end{eqnarray}}

\title{Detecting Information Relays in Deep Neural Networks}
\author[1,2,*]{Arend Hintze}
\author[2,3,4]{Christoph Adami}
\affil[1]{Department of MicroData Analytics, Dalarna University, Sweden}
\affil[2]{BEACON Center for the Study of Evolution in Action}
\affil[3]{Department of Microbiology and Molecular Genetics}
\affil[4]{Program in Evolution, Ecology, and Behavior, 

Michigan State University, United States of America}
\affil[*]{ahz@du.se}

\begin{document}
\maketitle

\abstract{
Deep learning of artificial neural networks (ANNs) is creating highly functional processes that are, unfortunately, nearly as hard to interpret as their biological counterparts. Identification of functional modules in natural brains plays an important role in cognitive and neuroscience alike, and can be carried out using a wide range of technologies such as fMRI,  EEG/ERP, MEG, or calcium imaging. However, we do not have such robust methods at our disposal when it comes to understanding functional modules in artificial neural networks. Ideally, understanding which parts of an artificial neural network perform what function might help us to address a number of vexing problems in ANN research, such as catastrophic forgetting and overfitting. Furthermore, revealing a network's modularity could improve our trust in them by making these black boxes more transparent.
Here, we introduce a new information-theoretic concept that proves useful in understanding and analyzing a network's functional modularity: the relay information $I_R$. The relay information measures how much information  groups of neurons that participate in a particular function (modules) relay from inputs to outputs. Combined with a greedy search algorithm, relay information can be used to {\em{identify}} computational modules in neural networks. We also show that the functionality of modules correlates with the amount of relay information they carry.
}

% Keywords
%\keyword{information theory, deep learning, relay} 

\captionsetup{width=\linewidth}
\section{Introduction}
Neural networks, be they natural or artificial deep-learned ones, notoriously are black boxes~\citep{castelvecchi2016can,adadi2018peeking}. To~understand how groups of neurons perform computations, to~obtain insight into the algorithms of the human mind, or~to be able to trust artificial systems, we need to make the network's processing more transparent. To~this end, various information-theoretic and other methods have been developed to shed light on the inner workings of neural networks. Transfer entropy~\citep{schreiber2000measuring} seeks to identify how much information is transferred from one node (or neuron) to the next, which in principle can detect causal links in a network~\citep{amblard2011directed} or be used to understand general properties about how information is distributed among nodes~\citep{tehrani2020can,hintze2020cryptic}. In~general, information theory can be used to make inferences in cognitive- and neuroscience~\citep{mcdonnell2011introductory,dimitrov2011information,timme2018tutorial}. Predictive information~\citep{bialek2001predictability,ay2008predictive} determines how much the outputs of a neural network depend on the inputs to the system or on hidden states. Integrated information~\citep{tononi2015integrated} quantifies how much a system combines inputs into a single experience and identifies the central component(s) in which that happens. Information theory is also used to determine cognitive control~\citep{fan2014information} and neural coding~\citep{borst1999information} in natural systems. Finally, information theory is used to characterize {\em {representations}}~\citep{marstaller2013evolution} that quantify how much (and where) information is stored about the~environment. 

 Despite the diverse range of applications of information theory to neuronal networks, the~question of which module or subset of nodes in an artificial neural network performs which function remains an open one. In~network neuroscience~\citep{sporns2022structure,hagmann2008mapping,SpornsBetzel2016}, this question is investigated using functional magnetic resonance imaging (fMRI)~\citep{logothetis2008we,he2009uncovering}, electroencephalography (EEG~\citep{thatcher2011neuropsychiatry}, magnetoencephalography (MEG) ~\citep{thatcher2011neuropsychiatry}, and~other physiological methods. fMRI specifically identifies functional cortical areas by the increase in oxygen consumption required to perform a particular task. However, an~fMRI detects two things at the same time: the activity of neurons involved in a specific task, and~the fact that they often form spatially associated clusters. As~a consequence, in~an fMRI analysis, functional and structural modularity coincide. In~deep convolutional neural networks, detecting modules based on their biological activity is obviously impossible. The~essential computation of the dot product of the state vector and weight matrix does not differ depending on how involved nodes and weights are in function. Furthermore, artificial neural networks do not display any structure beyond the order of layers. The~order of nodes in a layer is interchangeable, as~long as the associated weights change with them.

One approach to determine functional modularity in the context of ANNs is to determine the degree of modularity from the weights that connect nodes~\citep{shine2021nonlinear}, by~determining how compartmentalized information is~\citep{hintze2018structure,kirkpatrick2019role}, or~by performing a knockout analysis that allows tasks to be associated with the processing nodes of the neural network~\citep{cg2018effect}. However, results from such a knockout analysis are often not~conclusive.

Functional modularity in ANNs is interesting for another reason: it appears to affect a phenomenon known as {\em {catastrophic} forgetting} ~\citep{mccloskey1989catastrophic,french1999catastrophic}, where a network trained on one task can achieve high performance, but~catastrophically loses this performance when the network is sequentially trained on a new task. The~degree of modularity appears to be related to the method by which these networks are trained.
Using a genetic algorithm to modify weights (neuroevolution, see~\citep{stanley2019designing}) seems to produce modular structures automatically, as~was also observed in the evolution of metabolic networks~\citep{hintze2008evolution}. This modularity appears to protect ANNs from catastrophic forgetting~\citep{ellefsen2015neural}. Neural networks trained via backpropagation are unlikely to be modular since this training method recruits all weights into the task trained~\citep{hintze2021role}. Similarly, dropout regularization~\citep{hinton2012improving} is believed to cause all weights to be involved in solving a task (making the neural network more robust), which in turn prevents~overfitting.  

While many methods seek to prevent catastrophic forgetting~\citep{parisi2019continual}, such as Elastic Weight Consolidation (EWC) ~\citep{kirkpatrick2017overcoming}, algorithms such as LIME~\citep{ribeiro2016should}, and~even replay during sleep~\citep{Goldenetal2022},
it is still argued that catastrophic forgetting has not been solved~\citep{kemker2018measuring}. If~catastrophic forgetting is due to a lack of modularization of information, it becomes crucial to accurately measure this modularization to identify learning schemes that promote modules. 
The problem of identifying modules responsible for different functions
is further aggravated when information theory and perturbation analysis (via node knockout) disagree~\citep{bohm2022information,sella2022tracing}.

When identifying candidate neurons in hidden layers that might contain information about the inputs that are used in decision-making, perturbing those neurons by noise or knockout should disrupt function. Similarly, hidden nodes {\em not} containing information about inputs should, when perturbed in this manner, not alter the outputs encoding decisions. However, if~information is stored {\em redundantly}, perturbing only part of the redundant nodes will not necessarily disrupt function, even though they carry information. At~the same time, nodes without function or information can still accidentally perturb outputs when experiencing noise~\citep{bohm2022information,sella2022tracing}.

Here, we develop a new information-theoretic measure that quantifies how much information a set of nodes {\em {relays}} between inputs and outputs (relay information $I_R$). This measure can be applied to all combinations of neurons (sets) to identify which set of a given size contains the most information. While the number of sets of neurons is exponential in size, the~number of tests required to find the set with the largest amount of information can be significantly reduced by taking advantage of the fact that a smaller subset cannot have more information than its superset. Thus, this measure can be combined with a greedy search algorithm that identifies the relevant computational modules connecting the inputs to the outputs. We will demonstrate on a wide range of examples the function and applicability of this new method. Specifically, using a positive control in which the nodes relaying the information from inputs to outputs are known, we demonstrate that relay information indeed allows us to recover the relevant functional nodes. We compare this control to a regularly-trained neural network, and~show that perturbations on nodes carrying relay information cause failures in their predicted~functionality. 

\section{Methods}
\unskip

\subsection{Training Artificial Neural~Networks}
The neural networks used here are implemented using PyTorch~\citep{Paszkeetal2019} and trained on the MNIST handwritten numerals dataset~\citep{lecun1998gradient}. The~MNIST dataset consists of 60,000 training images and 10,000 test images of the ten numerals 0--9. Each grey-scale image has $28\times 28$ pixels with values normalized between $-1.0$ to $1.0$. Here, we use two different networks. The~\textit{full} network has 784 input nodes, followed by a layer of 20 hidden nodes with a standard summation aggregation function, and~a $\tanh$ threshold function. The~output layer needs 10 nodes to account for the ten possible numeral classes and uses the same aggregation and threshold function as the hidden layer. The~{\em composite} network is an aggregate of ten sub-networks each trained to recognize only a single number. In~each of the sub-networks, the~hidden layer has two nodes, with~a single node in the output~layer.

Networks are trained using the Adam optimizer~\citep{kingma2014adam} until they either reach a recognition accuracy of 95\% or else reach a given fixed number of training epochs. The~node in the output layer with the highest activation is used to indicate the network's prediction of the numeral depicted in the image (argmax function). 

\subsection{Composing an Artificial Neural Network from Specialized~Networks}
\begin{figure}[H]
\centering
		\includegraphics[width=5in]{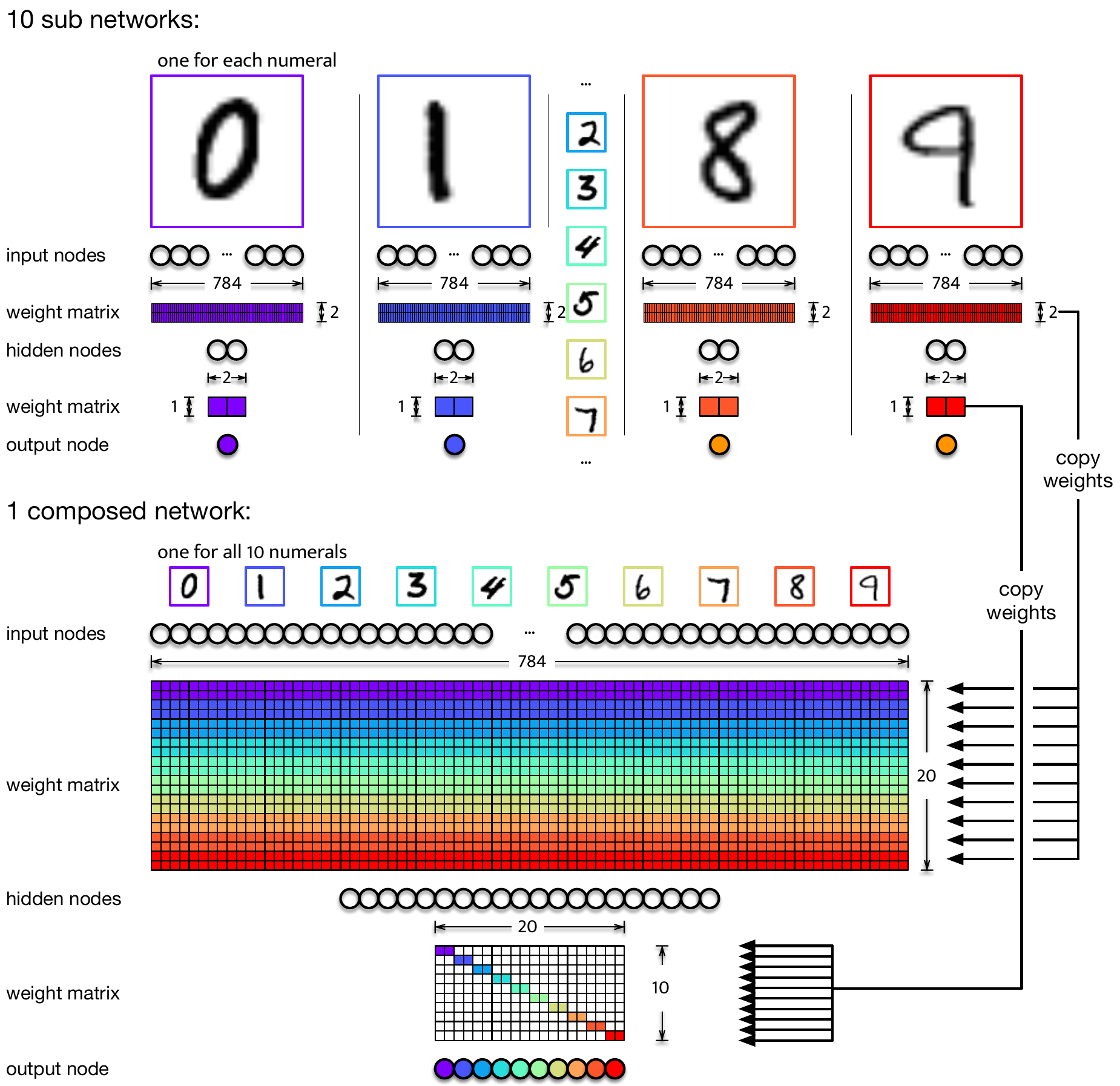}
	\caption{Illustration of the composite network. For~each of the ten numerals, an~independent neural network (sub-network) is trained to recognize a single numeral among the others. Each of those ten networks has 784 input nodes to receive data from the $28\times28$ pixel-wide MNIST images. Each hidden layer has two nodes followed by a single node at the output layer (top panel). The~composite network (bottom panel) is assembled from these ten subnetworks. Colors represent which weights in the combined weight matrix come from which corresponding sub-network. Weights shown as white remain $0.0$. Consequently, the~weight matrix connecting the hidden layer to the output layer is de facto~sparse.}
	\label{fig;compositionIllustration}
\end{figure}

In a typical ANN performing the MNIST classification task, all nodes of the hidden layer are involved in relaying the information from the input to the output layer: a phenomenon we previously termed {\em informational smearing}~\citep{hintze2018structure}, as the information is ``smeared'' over many neurons (as opposed to being localized to one or a few neurons). Our control network is constructed in such a manner that functionality is strictly distributed over very specific nodes. Specifically, we construct a network with 20 hidden nodes by aggregating ten sub-networks with two hidden nodes each. Each of the sub-networks is only trained to recognize a single numeral amongst the background of the other nine, using only two hidden nodes. By~combining these 10 sub-networks networks into the \textit{composite model}, we can create a control in which the relay neurons (the two hidden neurons in each of the sub-networks) are guaranteed to only relay information about a very specific function (see Figure~\ref{fig;compositionIllustration}). Note that 
those composite networks do not undergo further~training.

\subsection{Information-Theoretic Measure of Computational~Modules}
An artificial neural network can be viewed as an information-theoretic channel~\citep{shannon1948mathematical} that relays the information received at the input layer to the output layer while performing some computations along the way. To~measure the throughput of information, define the random variable $X_{\rm in}$ with ten states (one for each numeral) and Shannon entropy $H(X_{\rm in})$, while the outputs form a random variable  $X_{\rm out}$ with entropy $H(X_{\rm out})$. The~mutual information between both $I(X_{\rm in};X_{\rm out})$ (see~Equation~(\ref{equ;mutualInfo})) consequently measures how much the output symbol distribution is determined by the inputs (and vice~versa, as~this is an un-directed measurement):
\begin{equation}
    I(X_{\rm in};X_{\rm out})=H(X_{\rm in})+H(X_{\rm out})-H(X_{\rm out},X_{\rm in})\label{equ;mutualInfo}
\;.\end{equation}
Here, $H(X_{\rm out},X_{\rm in})$ stands for the joint entropy of the input and output~variables.

At the initialization of the network, weights are randomly seeded, giving rise to a network that randomly classifies images. In~this case, the~confusion matrix is relatively uniform and the conditional entropy $H(X_{\rm out}|X_{\rm in})=H(X_{\rm out},X_{\rm in})-H(X_{\rm in})\approx H(X_{\rm out})$, leading to low information $I(X_{\rm in};X_{\rm out})$. However, over~the course of training, the~prediction accuracy increases, leading ultimately to a strictly diagonal confusion matrix and a vanishing conditional entropy $H(X_{\rm out}|X_{\rm in})$, implying that every numeral is properly classified. In~this case, the~information channel has maximal information (information equals capacity) when measured over the training or test set. Note that, when we calculate the entropy of the inputs $H(X_{\rm in})$, we use only image labels (not all possible images).
%Fig. 1

We can view this joint channel as being composed of two sequential channels: one from the inputs to the hidden states, and~one from the hidden states to the outputs. The~information that the outputs receive is still determined by the inputs, but~now via the hidden variable $Y$. A~perfect channel can only exist if the hidden layer has sufficient bandwidth to transmit all of the entropy present at the inputs, that is, $H(Y)\geq H(X_{\rm in})$.

We can now write the information that flows from the inputs via the hidden states to the outputs in terms of the shared information between all three random variables
\begin{eqnarray}
    I(X_{\rm in};X_{\rm out};Y)&=&H(X_{\rm in})+H(X_{\rm out})+H(Y)\nonumber \\
    &-&H(X_{\rm in},X_{\rm out})-H(X_{\rm in},Y)-H(X_{\rm out},Y)\nonumber \\
    &+&H(X_{\rm in},X_{\rm out},Y)\;.\label{equ;inOutHidden}
\end{eqnarray}

\begin{figure}[H]
\centering
    \includegraphics[width=3in]{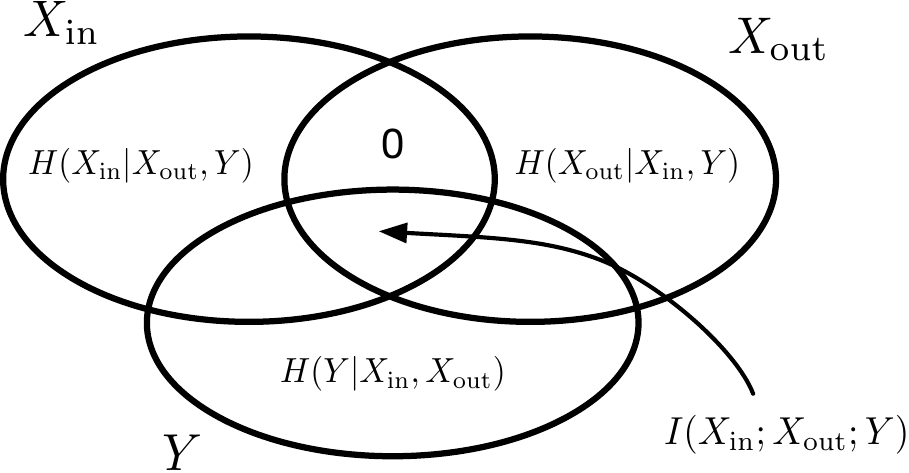}
    \caption{Entropy Venn diagram for the random variables $X_{\rm in}$, $X_{\rm out}$, and~$Y$. The~shared information between all three variables equals the information $I(X_{\rm in};X_{\rm out})$ because no information can flow from $X_{\rm in}$ to $X_{\rm out}$ without passing through $Y$.}
    \label{fig;venn1}
\end{figure}

Because information {\em must} pass through the hidden layer, this 
``triplet information'' must be equal to the information $I(X_{\rm in};X_{\rm out})$ (see Figure~\ref{fig;venn1}).
%Figure 2
However, in~general, not all of the nodes that comprise $Y$ carry information. Let us imagine, for~example, that the set of hidden nodes $Y$ is composed of a set $Y_R$ that shares information with $X_{\rm in}$ and $X_{\rm out}$, and~a set $Y_0$ that does not share this information, that is, $I(X_{\rm in};X_{\rm out};Y_0)=0$, with~$Y=Y_R\otimes Y_0$. We will discuss the algorithm to determine which neurons belong in the set of relay neurons $Y_R$ further below.

\begin{figure}[H]
\centering
    \includegraphics[width=4.5in]{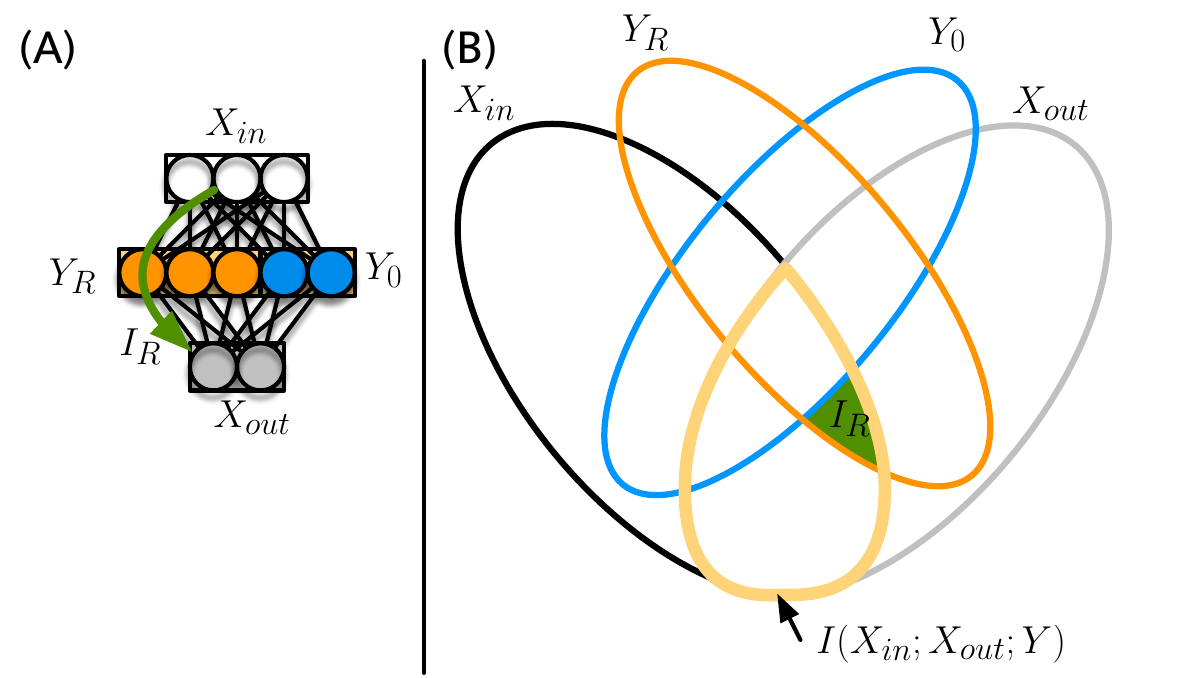}
    \caption{(\textbf{A}) Input/output structure of an ANN 
 with inputs $X_{\rm in}$, outputs $X_{\rm out}$, and~a hidden layer $Y=Y_R\otimes Y_0$. The~relay information passes from the inputs via the relay neurons $Y_R$ to the output (green arrow); (\textbf{B}) the entropic Venn diagram for the four variables $X_{\rm in}$, $X_{\rm out}$, $Y_R$, and~$Y_0$, with~ellipses quantifying the entropy of each of the variables colored according to (\textbf{A}). The~information shared between $X_{\rm in}$ and $X_{\rm out}$ is outlined in yellow. The~relay information Equation~(\ref{equ:relay}) is indicated by the green area.  \label{fig:Venn2}}
\end{figure}

The nodes that comprise $Y_0$ could, for~example, have zero-weight connections to the inputs, the~outputs, or~both. They are defined in such a way that none of the information $I(X_{\rm in};X_{\rm out}$) (area outlined in yellow in Figure~\ref{fig:Venn2}B) is shared with~them.
%Figure 3

We call the information that is relayed through the ``critical'' nodes that carry the information (the nodes in the set $Y_R$) the {\em relay information}. 
While we could define this information simply as the information shared between $X_{\rm in}$ that is also shared with the neurons identified to be in the set 
$\mathbb{Y_R}$ (see Section~\ref{ssaa}), it is important to deal with cases where neurons that are informationally inert (they do not read information from $X_{\rm in}$ nor write into $X_{\rm out}$) could nevertheless copy the state of a neuron that does relay information. In~the current construction, this does not appear to be very likely (or is at most a small effect). However, in~other architectures (such as recurrent neural networks, networks with multiple layers, probabilistic transfer functions, or~natural brains), such a phenomenon might be more common. As~discussed in Appendix~\ref{app1}, inert neurons that copy the state of relay neurons may be classified as belonging to the set $\mathbb{Y}_0$ (because removing them does not reduce the information carried by the set) yet show a nonvanishing $I(X_{\rm in};X_{\rm out};Y_0)$. In~order to eliminate such contributions, we measure the relay information {\em conditional} on the state of neurons in $Y_0$ that is
\begin{equation}
    I_R=H(X_{\rm in};X_{\rm out};Y_R|Y_0)\;, \label{equ:relay}
\end{equation}
which is indicated in the entropic Venn diagram in Figure~\ref{fig:Venn2} as the area colored in green. An~explicit expression for $I_R$ can be obtained simply by writing Equation~(\ref{equ;inOutHidden}) for $Y_R$ instead of $Y$, and~conditioning every term on $Y_0$.

We can also define a {\em particular relay information} (a relay information that pertains to any particular numeral class) by introducing the input-class random variable
\be
Z=Z_1\otimes Z_2\otimes \cdots \otimes Z_{10}\;.
\ee
Because we can decompose $X_{\rm out}$ in a similar manner
\be
X_{\rm out}=X_{\rm out}^{(1)}\otimes X_{\rm out}^{(2)}\otimes \cdots \otimes X_{\rm out}^{(10)}\;,
\ee
the relay information about numeral $i$ can then be written as
\be
I_R(i)=H(Z_i;X_{\rm out}^{(i)};Y_R|Y_0)\;.
\ee
This is the information that the critical relay nodes $Y_R$ are providing about numeral $i$.

The removal of hidden neurons that do not contribute to information transfer suggests a simple algorithm that identifies such neurons: start with the full set and remove neurons one by one, and~keep only those neurons that lead to a reduction of the information being relayed. However, this search is in reality more complex because neurons can carry redundant information. We discuss this algorithm in the following~section.

\subsection{Shrinking Subset Aggregation~Algorithm}
\label{ssaa}
In order to find the minimal subset of nodes $\mathbb{Y}_R$ that carry all of the information flowing from $X_{\rm in}$ to $X_{\rm out}$, we should in principle test all possible bi-partitions of neurons in~$Y$. 
Unfortunately, the~number of bi-partitions of a set is still exponential in the set size, so a complete enumeration can only be performed efficiently for small sets. However, it turns out that in most cases a greedy algorithm that removes nodes one by one will find the minimal set $\mathbb{Y}_R$ (see Appendix~\ref{app1}). 

We start with the largest partition in which all nodes belong to the set $\mathbb{Y}_R$, and~none to $\mathbb{Y}_0$. Now, all possible subsets in which a single node is moved from $\mathbb{Y}_R$ to $\mathbb{Y}_0$ can be tested. The~subset with the highest information (Equation (\ref{equ:relay})) is retained, and~the node with the lowest information contribution is permanently moved into subset $\mathbb{Y}_0$. This process is repeated until only one node is left in $\mathbb{Y}_R$. Over~the course of this procedure (assuming perfect estimation of entropy from sample data), the~set with the highest information for each set size should be identified (see Algorithm \ref{alg;ssaa}).

\begin{algorithm}
\caption{Shrinking Subset Aggregation~Algorithm.}\label{alg;ssaa}
\begin{algorithmic}
\Require $\mathbb{Y} = \{0, ..., n\}$\\

\State $\mathbb{Y}_0 \gets \emptyset$ 
\State $\mathbb{Y}_R \gets \mathbb{Y}$\\
\While{$\mathbb{Y}_R \neq \emptyset $}
    \For{$\forall a \in \mathbb{Y}_R$}
        \State $\mathbb{Y}_R' \gets \mathbb{Y}_R-{a}$
        \State $\mathbb{Y}_0' \gets \mathbb{Y}_0+{a}$
        \State $I_{a} \gets  I_R(X_{\rm in};X_{\rm out};\mathbb{Y}_R'|\mathbb{Y}_0)$ (see Equation~(\ref{equ:relay}))
    \EndFor
    \State $a \gets \{\mathbb{Y}_R; a = min(I_{a})\}$
    \State $\mathbb{Y}_R \gets \mathbb{Y}_R-{a}$
    \State $\mathbb{Y}_0 \gets \mathbb{Y}_0+{a}$
\EndWhile
\end{algorithmic}
\end{algorithm}
\unskip

As discussed in Appendix~\ref{app1}, this algorithm can sometimes fail to identify the correct minimal subset. First, estimates of entropies from finite ensembles can be inaccurate: these estimates are both noisy and biased (see, for~example,~\citep{Paninski2003}), leading to the removal of the wrong node from the set $\mathbb{Y}_R$.  Second, information can be stored redundantly. Imagine a network of ten nodes, with~three nodes forming the relay between inputs and outputs, while another set of two nodes is {\em redundant} with those other three nodes. The~greedy algorithm will work until all those five nodes are in the set $\mathbb{Y}_R$. Removing any of those nodes will not drop the information content of the larger set, since the information is fully and redundantly contained in both the set of three and the set of two. Thus, all five nodes appear equally {\em unimportant} to the algorithm, which can now not decide anymore which node to remove. It might remove one of the nodes in the set of three, leading to the set of two becoming the crucial computational module. Alternatively, removing a node from the smaller set promotes the larger set to become the crucial computational set. Either way, the~algorithm has a chance to fail to find a unique set because there could be~several.

One way to amend the algorithm would be to allow the process to dynamically branch.  In~case multiple nodes upon removal do not reduce the information retained in the remaining set $\mathbb{Y}_R$, all possible branches can be pursued. Such a fix will significantly increase the computational time. However, as~we do not expect the occurrence of redundant sets to be a prominent feature of many networks, we have not explored this alternative algorithm~further.

\subsection{Knockout~Analysis}
To test the informational relevance of individual nodes of the hidden layer, we can perform ``knockout'' experiments. While a knockout in a biological context is defined as the disabling of a component, it is less obvious how to perform such an operation in the context of a neural network. One option would be to replace a neuron's activation level by a random number, which still leaves the freedom to choose a distribution and a range. Furthermore, these random values still propagate through the network, which implies that such a knocked-out neuron is not disabled. Keeping an activation level constant (independent of the inputs) can also have undesirable effects. Imagine that a neuron's activation level is constant, say $1.0$ or $-1.0$, independently of the input images. This value would be included in all subsequent aggregation functions affecting the final output of the network. Attempting to knock out this node by forcing it to $-1.0$ or $1.0$ can now have two different effects. If~the node is already a constant $1.0$, knocking it out by forcing it to be a constant $1.0$ would suggest that this node has no function, since such a knockout would not change any output. Setting it to $-1.0$ might have significant effects, but~would on the other hand leave a node that should be at $-1.0$ unaffected. Here, to~``knock out'' a node (to render it non-functional) in the hidden layer, we force it to take on a value of $0.0$ during the forward pass. At~the same time, all weights of the first layer leading to such a node are set to $0.0$, as~are all weights of the second layer that are affected by that node. Alternatively, the~values of the nodes to be knocked out in the hidden layer could have been forced to $0.0$ when the forward pass reaches the hidden layer. These methods are equivalent. While this form of knockout can also have undesirable consequences, the~effect is likely closest to an actual removal of the node by eliminating it from the network, and~shrinking the weight matrices~accordingly.

\subsection{Coarse-Graining Continuous~Variables}
The computations performed by the neural network use continuous inputs, and~due to the $\tanh$-like threshold function, the~activation levels of neurons in the hidden layer are confined to the interval $[-1,1]$. While entropies can be computed on continuous variables (so-called differential entropies, see~\citep{shannon1948mathematical}), we use discrete entropies here, which require a discretization of the continuous values. In~particular, we are coarse-graining those entropies by mapping all continuous values to the binary categories $0$ and $1$. We previously used the median value of a neuron's excitation level as the threshold for the bin~\citep{bohm2022understanding}. Instead, here the hidden-state values are clustered using a $k$-means clustering algorithm with $k=2$. Using the median for coarse-graining ensures that the resulting distribution has maximal entropy because each bin is guaranteed to receive half of the values. However, we found a maximum-entropy assumption for a neuron to be inappropriate in most cases. Using a $k$-means clustering algorithm to distribute values into bins gives a better approximation of the relative entropy between~states.

Coarse-graining also reduces the size of the state space that is being sampled. Using $k=2$ to coarse-grain the hidden states implies that there are at most  $k^N$ possible states, which (with $N=20$) is a state space that is in danger of being significantly undersampled with a limited number of images (MNIST has at most 70,000 if test and training data are combined). As~the entropy of this hidden space is $N\log_2 k$ bits, an~input sample with $\log(60,000)\approx 15.87$ bits would be insufficient to adequately sample the random variables $X_{\rm in}$, $X_{\rm out}$, $Y_{R}$, or~$Y_{0}$ even for $k=2$. However, as~discussed in Appendix~\ref{app2}, because~the input entropy is much smaller ($\log 10\approx 3.32$ bits), estimation errors are small, and~the likelihood that nodes are accidentally removed from $Y_R$ due to poor sampling is~small.

\subsection{Aggregated Relay~Information}
The greedy algorithm identifies a sequence of sets of nodes that continuously shrink because it is always the node contributing the least to $I_R$ that is removed next. Consequently, every time a node is removed, we can also quantify the loss of information for that particular node $n$ as the difference in $I_R$ between the larger set containing the node ($\mathbb{Y}_R \cup n$) and smaller set 
 without it ($\mathbb{Y}_R$):
\begin{equation}
     \Delta I(n)=I_R(\mathbb{Y}_R\cup n)-I_R(\mathbb{Y}_R)\;. \label{equ:infoDrop}
\end{equation}

Interestingly, when this process arrives at a set of nodes that taken together is essential in relaying the information, it can happen that the removal of {\em any} of the nodes of the set causes the remaining neurons to have $I_R=0$. 
Information in such an essential set can be seen to be {\em encrypted}, to~the point where no node can be removed without losing all of the information~\citep{bohm2022information}.
However, this creates a situation in which the last nodes, when removed, appear to not contribute any information, even though they are essential.
Thus, we quantify the amount that each node contributes to the relay information in terms of the sum of all $\Delta I(n)$ over all previously removed nodes as
\begin{equation}
    I_A(n)=\sum_{i=1}^{n} \Delta I(i)\;.
    \label{equ:aggregateRelayInformation}
\end{equation}
 
Using the information loss due to the removal of a node from the essential set, we can also quantify the {\em essentiality} of a neuron in terms of the loss of information the removal of node $n$ causes when it is 
removed from the remaining set of nodes. The~essentiality of a single node can be computed using Equation~(\ref{equ:infoDrop}) where $n$ is the node being removed from the full set of nodes. Thus, if~a neuron is meaningless or redundant, its essentiality $\Delta I(n)$ will~vanish.

\section{Results}
\unskip

\subsection{Identification of Information~Relays}
To determine if the proposed metric and optimization method correctly identifies the nodes that relay information from the outputs to the inputs, we trained two kinds of networks. A~standard ANN with 20 hidden nodes was trained to correctly identify all ten numerals. As~a control, ten sub-networks with two hidden nodes were trained on a single numeral each. From~the ten smaller networks, a~full network was composed (see Figure~\ref{fig;compositionIllustration}) that can perform the same task as the network trained on all numerals at the same time.

\begin{figure}[H]
\centering
		\includegraphics[width=5in]{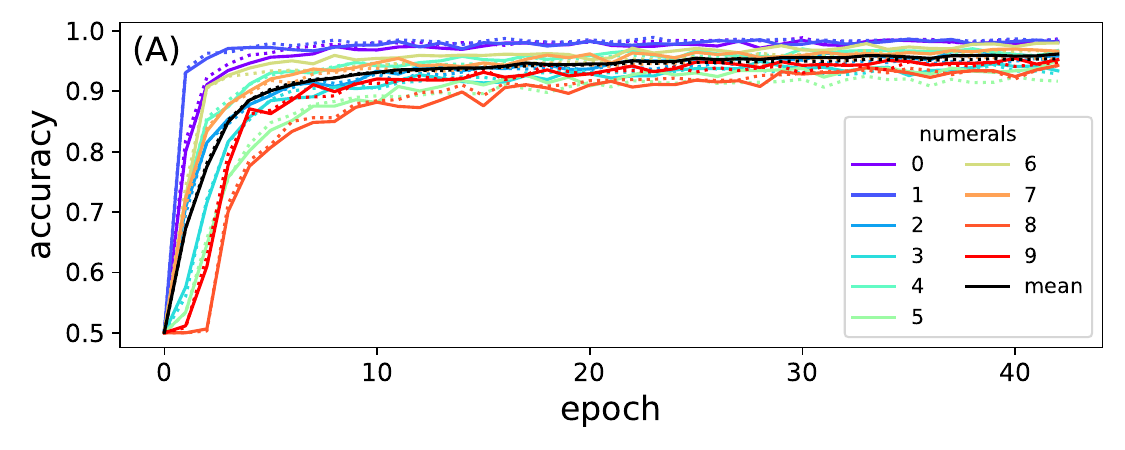}
		\includegraphics[width=5in]{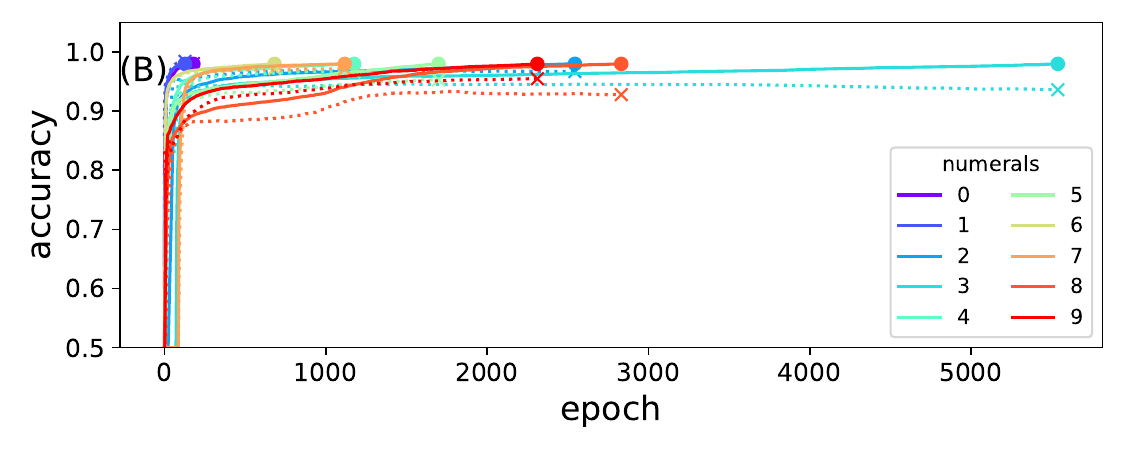}
	\caption{Training accuracy as a function of training epoch. (\textbf{A}) full model (top panel). The~accuracy to predict each numeral is indicated with lines of different colors (see legend). Accuracy on the training set is shown as solid lines while accuracy on the test is indicated by dotted lines. The~average performance classifying all numbers is shown in black; (\textbf{B}) accuracy of each of the ten sub-network models used to create the composite model as a function of training epoch. Colors indicate the accuracy for detecting an individual numeral. The~endpoint of the training is highlighted with a dot; the same time point but using test data is indicated by an x. Training other networks had marginally different outcomes (data not shown).  } 
	\label{fig;training}
\end{figure}

Figure~\ref{fig;training} shows the mean accuracy of recognizing each of the different digits as a function of training epoch, for~the full as well as the composite network. 
Note that the full network only needed 43 epochs to reach 96\% accuracy, while the training of the smaller models took significantly longer. The~full model was trained until it reached an accuracy of 0.96; the smaller models were trained until they reach an accuracy of 0.98.
Smaller networks could easily be trained to achieve this high 98\% accuracy while training the full network is usually limited to 96\%. In~order to observe networks performing as optimally as possible, and~to maximize the information between inputs and outputs, networks were trained until they reached those practical limits~\citep{chapman2013evolution}.
 
%Figure 4

Because in the composite network the two hidden neurons of each sub-network are guaranteed to serve as relays for the relevant information, we can use this network as a positive control to test whether our algorithm correctly detects relay information, and~whether neurons carrying non-overlapping information (each of the hidden neuron sets only carries the information about one specific numeral) are either more or less vulnerable to knockout. This does not imply that the hidden neurons that pertain to a particular numeral cannot relay information about another numeral. After~all, hidden nodes trained to recognize the numeral 1, for~example, might still correlate with nodes trained to recognize numeral 7 due to the similarity between those~images.

In order to test whether the greedy algorithm finds the correct minimal informative subset in the full model, we performed an exhaustive search of all $2^N-1$ (with $N=20$) bi-partitions of the hidden nodes $Y$ to find the minimal set $Y_R$. We then compared the result of the exhaustive search with the candidate set resulting from the shrinking subset aggregation algorithm. This un-branched version of the algorithm only needs $\frac{N(N+1)}{2}$ computations, reducing the computational complexity from exponential to~quadratic.

\begin{figure}[H]
\centering
        \textbf{Full model}\par\medskip
		\includegraphics[width=4.5in]{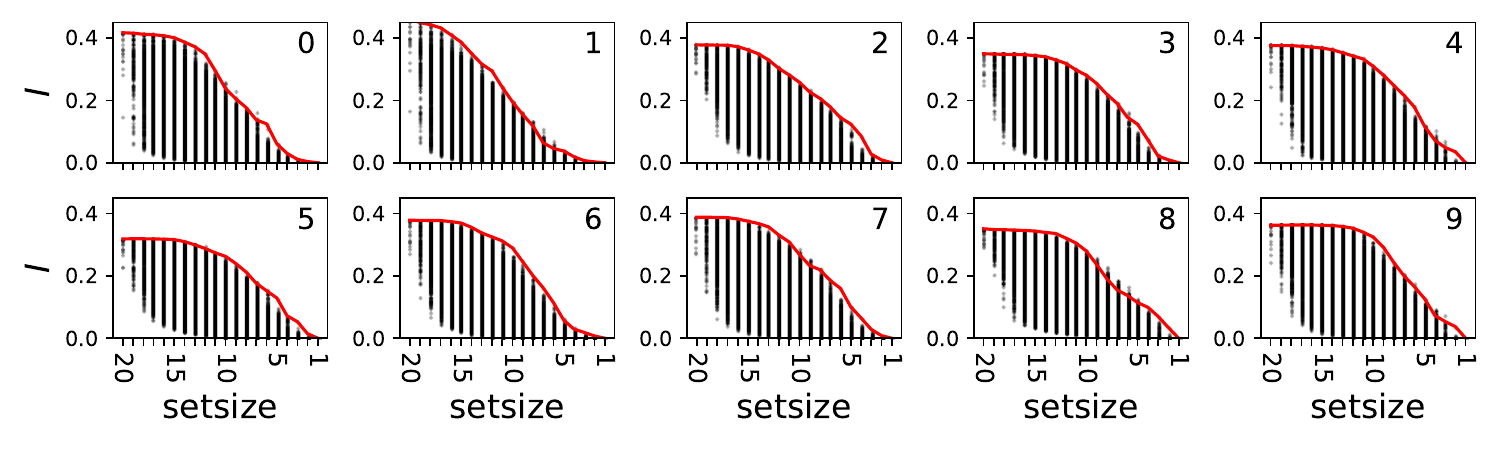}\\
        \textbf{Composite model}\par\medskip
		\includegraphics[width=4.5in]{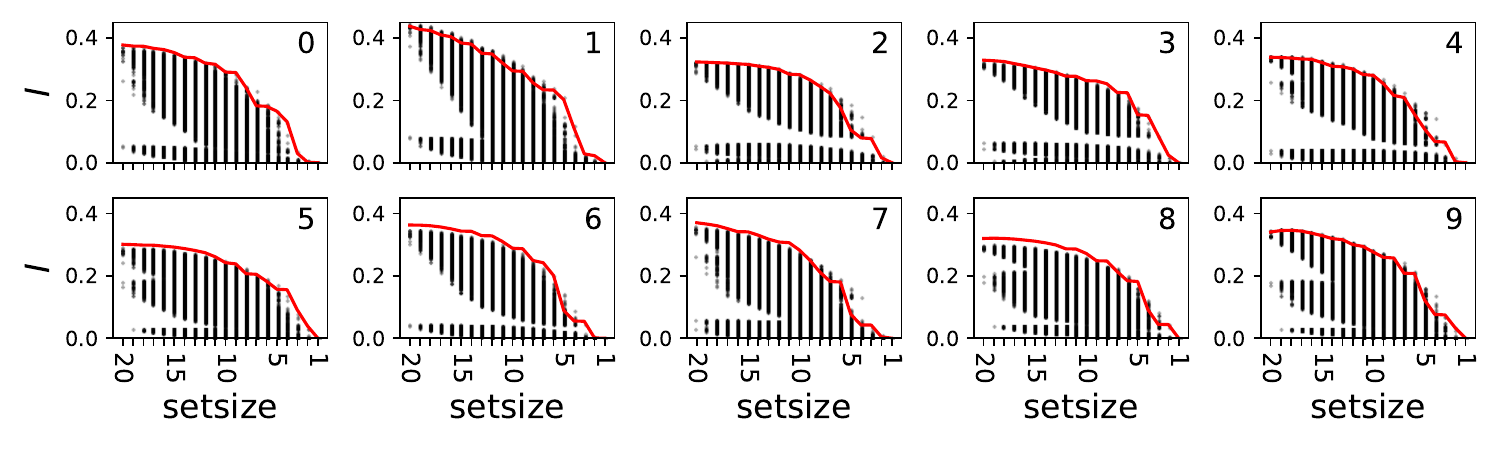}
	\caption{Particular relay information about each numeral for all possible bi-partitions (black dots) as a function of the set sizes $|\mathbb{Y}_R|$. The~top ten panels show particular relay information for the full model, while the bottom ten panels show the same for the composite model. Each panel shows the relay information about a different numeral in the MNIST task, indicated by the index of the panel. The~red line corresponds to the set identified by the shrinking subset aggregation algorithm. Fewer than 0.9\% of all subsets have a higher information content than the one identified by the~algorithm.}
	\label{fig;setAnalysis}
\end{figure}

Figure~\ref{fig;setAnalysis} shows that different partitions relay very different amounts of information about the particular output.
In general, the~larger the set $\mathbb{Y}_R$, the~more information it represents, but~we also see that the highest information found within sets of a particular size is always higher than the maximal information found amongst all sets that are smaller 
(as proved in Appendix~\ref{app1}, with~the caveat of redundant sets). The~shrinking subset aggregation algorithm exploits this observation of smaller sets always having less information than their larger superset
and should thus be capable of identifying the subsets $\mathbb{Y}_R$ (and consequently also $\mathbb{Y}_0$) with the highest information content for all sets of the same size, but~without the complete enumeration of all possible sets. We find that fewer than 0.9\% of the correct sets have equal or more information than the set identified by the greedy algorithm. As~discussed earlier, the~failure of the greedy algorithm to correctly identify the most informative set can be attributed to noise in the entropy estimate due to the finite sample size, as~well as to the presence of redundant sets with identical information~content. 

\begin{figure}[H]
		\includegraphics[width=5.5in]{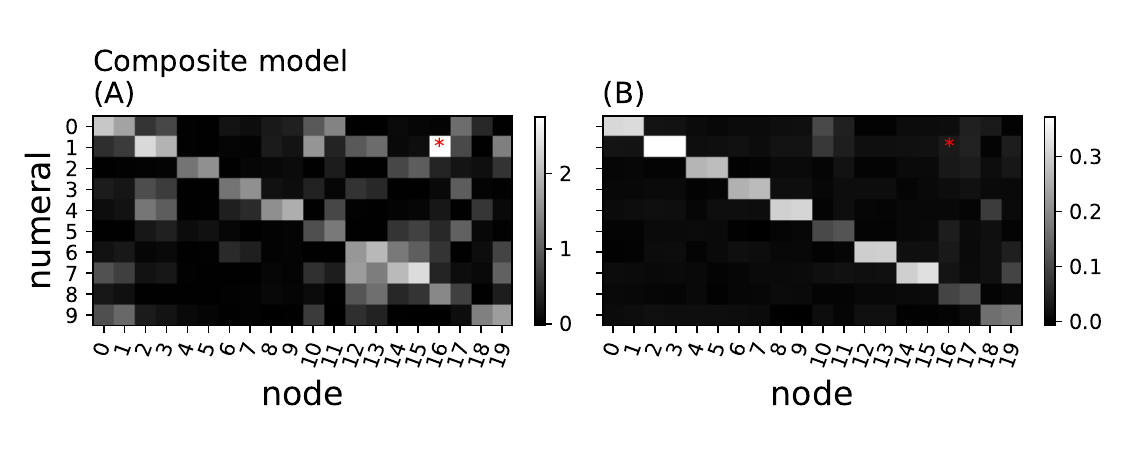}
 	\caption{Aggregated relay information and essentiality in the composite model. (\textbf{A}) aggregated particular information loss $\Delta I_R(n)$ (Equation~(\ref{equ:aggregateRelayInformation})) for all 20 nodes in the hidden layer ($x$-axis) and the ten different numeral classes ($y$-axis) shown in grayscale (brighter shades indicate higher loss of information); (\textbf{B}) node essentiality (Equation~\ref{equ:infoDrop})) for each hidden neuron and numeral. Bright squares indicate essential nodes, while black squares would indicate redundant or meaningless nodes. The~red dot (node 16, numeral 1) points to a neuron that appears to relay information (\textbf{A}) but is entirely redundant and non-essential (red dot in (\textbf{B})).} 
	\label{fig;nodeResultComposed}
\end{figure}

We now investigate whether the greedy algorithm properly identifies the relevant subsets that are critical in relaying the information from inputs to outputs that is, whether the information they carry is indeed used to predict the depicted numeral. We define the {\em importance} of a node as the sum of all information loss that this node conveyed before it was removed (aggregated relay information, see Methods). We also define the {\em essentiality} of node $n$ as the amount of relay information lost when moving that node from the minimal set $Y_{R}$ to $Y_{0}$ (see Equation~(\ref{equ:infoDrop})). Because~this measure of essentiality only considers the effect of removing single nodes, it can be inaccurate if a node is essential only when another node (for example a redundant one) is also removed. However, since the relays in the composite network are so small (two nodes) removing any one of them causes a large drop of information. This can be seen in Figure~\ref{fig;nodeResultComposed}B, where nodes identified as relays are also highly~essential.

Figure~\ref{fig;nodeResultComposed}A shows that both the importance analysis (via the aggregated particular relay information) and the essentiality analysis correctly identify the nodes that relay the information from inputs to outputs in the composite model. Aside from the sampling noise, each pair of hidden nodes that were trained to be the relays are correctly identified as highly informative (see Figure~\ref{fig;nodeResultComposed}). \vspace{-6pt} 

\begin{figure}[H]
\centering	
\includegraphics[width=5.5in]{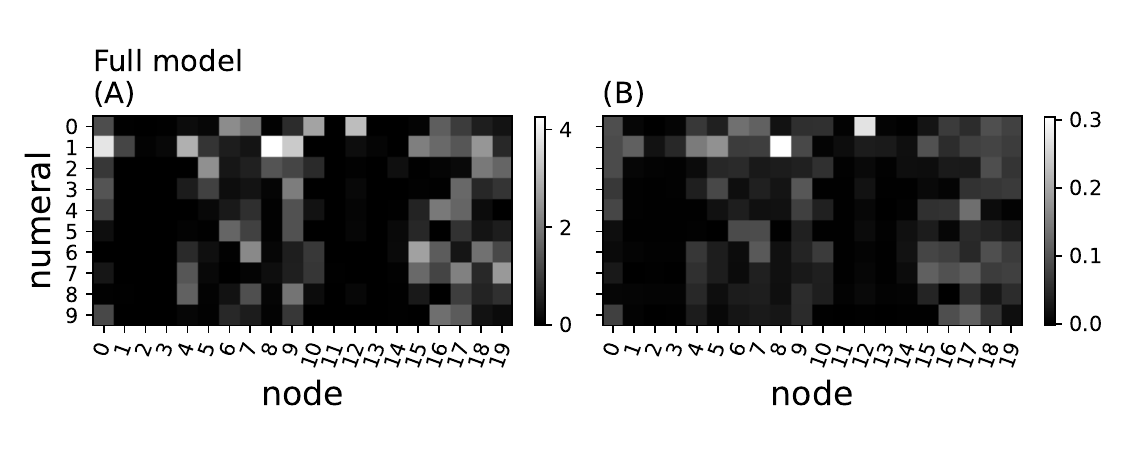}
\caption{Aggregated relay information and essentiality in the full model. (\textbf{A}) aggregated relay information for each node and every numeral class for the full network; (\textbf{B}) essentiality. Methods, axes, and~grayscales as in Figure~\ref{fig;nodeResultComposed}. } 
	\label{fig;nodeResultFull}
\end{figure}

Training the full network via backpropagation is not expected to create modules of hidden nodes that each only relay information about one specific numeral. Indeed, we find information to be relayed in an unstructured fashion in this network (see Figure~\ref{fig;nodeResultFull}A). Interestingly, nodes that are positively identified as relays are not necessarily essential, suggesting that many nodes contain redundant information (see Figure~\ref{fig;nodeResultFull}B). This further supports our previous findings that backpropagation smears or distributes function across all nodes, rather than isolating functions into structured modules~\citep{hintze2018structure,kirkpatrick2019role}. The~results from Figure~\ref{fig;nodeResultFull}B also suggest that using the essentiality of single nodes does not properly identify the informational structure of the~network.

\subsection{Information Relays Are Critical for the Function of the Neural~Network}
\begin{figure}[H]
\centering
		\includegraphics[width=5.0in]{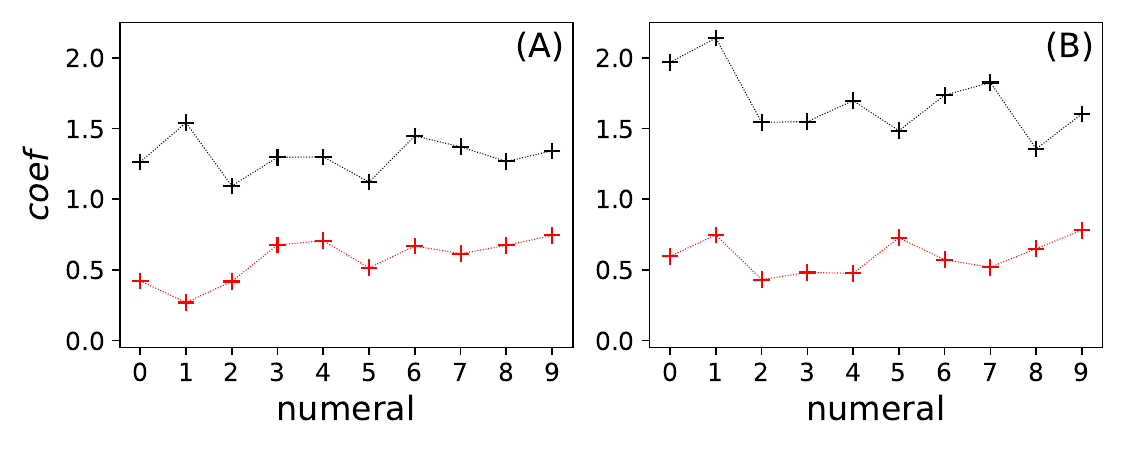}
	\caption{Regression coefficients of the multiple linear regression analysis between knockout effect $K$ and set size $|\mathbb{Y}_R|$ (red crosses), and~knockout effect $K$ and particular relay information $I_R(i)$ (black crosses), as~a function of numeral $i$. Lines are meant to guide the eye. (\textbf{A}) full model; (\textbf{B}) composite model.
	\label{fig;multipleRegression}} 
\end{figure}

To verify that the sets $\mathbb{Y}_R$ with high information are indeed relaying information from the inputs to the outputs, we can study the effect of knockouts on those nodes. Because~we expect a correlation between knockout effect size (the sensitivity of the node to perturbation) and the size of the informative set, care must be taken when interpreting the correlation between relay information and knockout effect size (sensitivity). Smaller sets can relay less information and have a potentially smaller effect when knocked out compared to larger sets. Thus, set size confounds the correlation between knockout effect and the amount of information relayed by the same set. We performed a multiple linear regression to test how much the knockout effect (treated as the dependent variable) is explained either by the set size or the amount of information relayed (independent variable). Figure~\ref{fig;multipleRegression} shows the regression coefficients of that~analysis.
\vspace{-6pt} 
%Fig. 6

Relay information explains at least 75\% ($r^2>0.75$) of the variance of the knockout effect for the composite model and at least 45\% ($r^2>0.45$) of the variance of the knockout effect for the full model. 
We can thus conclude that, when assuming a linear relationship between either set size or relay information and knockout effect, the~influence of relay information on knockout effect is significantly stronger than the influence of set size ($F>1.5\times 10^5$ in an F-test).

Figure~\ref{fig;multipleRegression} shows that the knockout effect is better explained by the amount of particular relay information about that node than the set size $|\mathbb{Y}_R|$. This shows also that, as~expected, set size is indeed confounding this relation. We further find that in the composite network the relationship between particular relay information and knockout effect is stronger compared to the full network. The~weaker relation between knockout effect and relay information is most likely due to the information being distributed more broadly over many nodes, compared to the composite model where the information is forced to reside in only two relay~nodes. 

%Figure 8
\unskip

\section{Discussion}
We introduced a new information-theoretic concept that we believe will prove to be useful in the analysis of information flow in natural and artificial brains: the ``relay information''. Relay information quantifies the amount of information within a set of nodes inside a communication channel that passes through this set, and~not through other nodes within the same channel. The~particular relay information can be used to identify which nodes in a hidden layer of a neural network are responsible for what particular classification function. We constructed a greedy algorithm that identifies the minimal informative set of nodes that carry the particular relay information, and~tested it on the MNIST hand-written numeral classification task using a regular neural network, as~well as a control in a network in which we know---by construction---the function of each hidden node (see Figure~\ref{fig;compositionIllustration}). We further showed via a knockout analysis that the sets of neurons identified as carrying the relay information indeed are functional because knocking out those nodes abrogates the classification accuracy for that particular~numeral. 

The identification of information relays, and~thus discovering the computational modules that relay information, can only be a first step in a more comprehensive analysis of brain function. Here, we focused on testing the method, and~showed using a positive control (the composite network) that the identified relay sets are indeed correlated to function. We also found that the full network, trained on all image classes at the same time, does not display a well-differentiated modular structure. Instead, information is distributed haphazardly across the network, and~if we were to identify functional modules, they would be highly overlapping. In~other words, the~ANNs that we trained here do not seem to have a modular structure in information~space.

Because a defined, modular, informational structure appears to be key to understanding a number of key properties of neural networks (such as catastrophic forgetting~\citep{mccloskey1989catastrophic,kirkpatrick2017overcoming,kemker2018measuring} or learning~\citep{ellefsen2015neural}), understanding what design decisions give rise to more (or less) modular networks is an important first step.  We are now better equipped to study the role of information smearing and modularity in its effect on fooling, generalization, catastrophic forgetting, or~latent space variables, and~look forward to exploring these topics in the~future.

The concepts and methods we introduced are general and can be applied to any task where a network (be it neural or genetic) performs its function by taking inputs, computing outputs, and~then using those outputs for prediction. In~the case of a natural neural network, however, because~it is continuously processing, an~additional temporal binning has to be performed. This, and~measuring the states of \textit{all} neurons in the first place, will make applying the concept of relay information challenging, to~say the least. In~the future, it would be interesting to study if this method also applies to, for~example, time series classification, recurrent neural networks, convolutional layers, or~even generative~tasks. 

Another concern is the scaling of the computational complexity of the algorithm to detect information relays with the number of nodes in the hidden layer. Currently, using the greedy algorithm and all 60,000 training images from the MNIST data set, and~applying it to a full network with 20 hidden nodes, takes about 30 s on a 3.5 Ghz desktop computer (for all 10 numerals together). Performing the same analysis but computing the exact critical set (testing all $2^N$ sets) takes about 24 h on the same hardware. Because~the complexity of the greedy algorithm has a computational complexity of $O(N(N-1))$ and the full enumeration has a computational complexity of $O(2^N)$, we can conjecture that a network of 1000 nodes can be analyzed within the same 24 h needed for a network of size $N=20$.

In this work, we only studied one particular optimizer to train the neural network (Adam), one loss function (mean squared error), and~the threshold functions hyperbolic tangent and argmax. We conjecture that our method applies to all other variances of deep learning. However, we also conjecture that the way in which information is distributed across the network will depend on the method and parameters of the optimization procedure, and~we will test this dependence in future work. Finally, by~testing different coarse-grainings of neuronal firing levels, the~method should be able identify relay neurons and thus functional modules in biological brains, and~thus help in studying information flow in functioning~brains.

In this work, we found that the greedy algorithm correctly identifies the minimal informative set in almost all cases. However, we expect that the failure rate depends on the task being studied, the~data set size, as~well as the amount of redundancy among neurons. In~networks with significant redundancy, we can imagine that the algorithm fails significantly more often, in~which case a branching algorithm may have to be designed, which would carry a significant complexity~cost.

%%%%%%%%%%%%%%%%%%%%%%%%%%%%%%%%%%%%%%%%%%
\vskip 0.25cm
\noindent{\bf Author contributions}\\
A.H. implemented all computational analysis and methods, A.H. and C.A. designed the experiments and devised the new methods, and~A.H. and C.A. wrote the~manuscript.  All authors have read and agreed to the published version of the manuscript.\\
\vskip 0.25cm
\noindent{\bf Funding information}\\ 
This research was supported by the Uppsala Multidisciplinary Center for Advanced Computational Science SNIC 2020-15-48, and~the National Science Foundation No. DBI-0939454 BEACON Center for the Study of Evolution in~Action.\\
\vskip 0.25cm
\noindent{\bf Data availability}\\
Code for the computational experiments and the data analysis can be found at\\
 \url{https://github.com/Hintzelab/Entropy-Detecting-Information-Relays-in-Deep-Neural-Networks} DOI:10.5281/zenodo.7660142\\
\vskip 0.25cm
\noindent{\bf Acknowledgements}\\
We thank Clifford Bohm for extensive discussions. This research was supported by Uppsala Multidisciplinary Center for Advanced Computational Science SNIC 2020-15-48, and the National Science Foundation No. DBI-0939454 BEACON Center for the Study of Evolution in Action.

%%%%%%%%%%%%%%%%%%%%%%%%%%%%%%%%%%%%%%%%%%
\appendix

\section[\appendixname~\thesection]{Proof of non-deceptive removal of nodes}

\label{app1}
Here, we show that, as~long as information is not redundantly encoded, it is possible to remove nodes one by one in a greedy fashion so that the minimal information reduction by single-node removal is not deceptive. In~a deceptive removal, removing a pair of nodes reduces the information by a smaller amount than each of the individuals would have~removed.

Say that the information to predict feature $X_{\rm out}$ is stored in $n$ variables $Y_1\cdots Y_n$. This information is
\be
I(X_{\rm out};Y_1\cdots Y_n)\;.
\ee
The general rule to study node removal is the identity
\be
I(X_{\rm out};Y_1\cdots Y_n)=I(X;Y_1\cdots Y_{n-1})+H(X_{\rm out};Y_n|Y_1\cdots Y_{n-1})\;. \label{rule}
\ee
We can easily convince ourselves of this, by~imagining $Y=Y_1\cdots Y_{n-1}$ and $Y_n=Z$; then, this equation is just
\be
I(X;YZ)=I(X;Y)+H(X;Z|Y)\;, \label{rule1}
\ee
which is easily seen by writing the Venn diagram between $X$, $Y$, and~$Z$.

Let us study the simplest case of three nodes. The~three possible information reductions are
\be
\Delta I_1&=&H(X_{\rm out};Y_1|Y_2Y_3)\;,\\
\Delta I_2&=&H(X_{\rm out};Y_2|Y_1Y_3)\;,\\
\Delta I_3&=&H(X_{\rm out};Y_3|Y_1Y_2)\;.
\ee
We will now prove that, if~$\Delta I_3<\Delta I_1$ and {\em at the same time} $\Delta I_3<\Delta I_2$ (implying that node 3 should be removed first), then it is {\em not} possible that the information reduction due to the removal of nodes 1 and 2 at the same time is smaller than the information loss coming from the removal of node 3 (i.e., that $\Delta I_{12}<\Delta I_3$ is impossible). 
If the latter was possible, we should have removed nodes 1 and 2 at the same time instead of node 3 (making the removal of node 3 deceptive). 

Let us first write down $\Delta I_{12}$. Since
\be
H(X_{\rm out};Y_1Y_2Y_3)=I(X_{\rm out};Y_3)+H(X_{\rm out};Y_1Y_2|Y_3),\
\ee
we know that
\be
\Delta I_{12}=H(X_{\rm out};Y_1Y_2|Y_3)\;.
\ee
We can rewrite this as
\be
H(X_{\rm out};Y_1Y_2|Y_3)=H(X_{\rm out};Y_2|Y_3)+H(X_{\rm out};Y_1|Y_2Y_3)\;. \label{eq}
\ee
This is the same rule as (\ref{rule1}), just conditioned on a variable (all information-theoretic equalities remain true if the left-hand side and the right-hand side are conditioned on the same variable).
Equation~(\ref{eq}) implies that
\be
\Delta I_{12}=H(X_{\rm out};Y_2|Y_3)+\Delta I_1\;.
\ee
Now since $H(X_{\rm out};Y_2|Y_3)\geq0$, we know that $\Delta I_{12}\geq \Delta I_1$. However, since $\Delta I_1>\Delta I_3$ by assumption, it follows immediately that
\be
\Delta I_{12}>\Delta I_3
\ee
contradicting the claim that it is possible that $\Delta I_{12}<\Delta I_3$.

Clearly, the~same argument will apply if we ask whether larger groups are removed first: they can never remove less information than the smallest information removed by a single node in that~group. 

If information is redundantly encoded, the~greedy algorithm can fail. Suppose two nodes are copies of each other $Y_1=Y_2$, making them perfectly correlated: they carry the same exact information about $X_{\rm out}$. In~that case, we can remove any of the two nodes, and~it will not change the information, that is, $\Delta I_1=\Delta I_2=0$:
\be
I(X_{\rm out};Y_1Y_2Y_3)=I(X_{\rm out};Y_1Y_3)=I(X_{\rm out};Y_2Y_3)\;.
\ee
However, once we removed one (say we removed $Y_1$), then removing $Y_2$ potentially removes information, as~\be
I(X_{\rm out};Y_2Y_3)=I(X_{\rm out};Y_3)+H(X_{\rm out};Y_2|Y_3)\;.
\ee
Now that $Y_1$ is removed, the~redundancy is gone, and~$H(X_{\rm out};Y_2|Y_3)$ could be large even though $\Delta I_1=H(X_{\rm out};Y_1|Y_2Y_3)=0$. This failure of the greedy algorithm is at the origin of the discrepancies between the true minimal set of nodes (obtained by exhaustive enumeration) and the set identified by the greedy algorithm, but~the failure is clearly rare.

\section[\appendixname~\thesection]{Sampling Large State Spaces}
\label{app2}
In this work, we identify relay information by calculating shared conditional entropies such as those shown in Equation~(\ref{equ;mutualInfo}). Calculating those entropies, in~turn, relies on estimating the entropy of the hidden layer neurons $H(Y)$, which can become cumbersome if the number of neurons in the hidden layer is large. To~calculate a quantity such as $I(X_{\rm in};X_{\rm out};Y)$ (the center of the diagram in Figure~\ref{fig;venn1}), we must estimate probabilities such as $p(y)$, the~probability to find the hidden layer in any of its $2^{20}$ states, assuming a binary state for each node after binning. To~obtain an accurate maximum-likelihood estimate of $p(y)$, the~sample size has to be large. In~order to estimate entropies, for~example, a~good rule of thumb is that the finite sample size bias~\citep{Basharin1959} 
\be
\Delta H=\frac{M-1}{2N\ln k}<<1\;, \label{condition}
\ee
where $M$ is the number of states in $Y$, $N$ is the sample size, and~$k$ is the dimension of the alphabet ($k=2$ for a binary random variable). Evidently, for~$M=2^{20}$ and $N=60,000$, this condition is not fulfilled. 
However, because~the output entropy $H(X_{\rm out})$ is constrained to be $\log_2(10)\approx 3.32$ bits, a~trained ANN should never end up with a hidden layer entropy near $\log M$, but~rather have an entropy comparable to that of the output state. In~this way, the~channel loss $H(Y|X_{\rm out})$ is~minimized. 

To test whether hidden layer entropy estimates are adequately sampled, we measured the entropy of the hidden layer when sampling over the entire image space. If~$Y$ was uniformly distributed, we would expect $H(Y)\approx \log(N)\approx 15.86$. Instead, we found the actual entropy $H_{\rm act}(Y)\approx 3.319$, indicating that the hidden layer probability distribution is highly skewed. In~this manner, the~effective number of states 
$M_{\rm eff}=2^{H_{\rm act}(Y)}\approx 9.98$, which easily satisfies condition (\ref{condition}).

\begin{figure}[H]
		\includegraphics[width=5.0in]{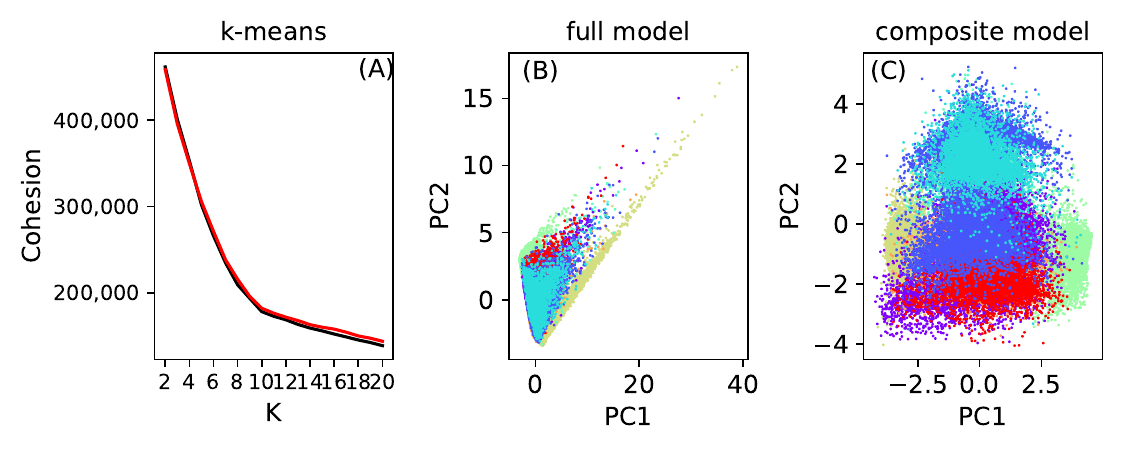}
	\caption{(\textbf{A}) $k$-means clustering performed on
  the hidden states of the full (black) and composite (red) neural network. The~elbow method deployed here tests the cohesion (sum of squared distances between each point and its associated centroid, $y$-axis) against the number of clusters ($k$, $x$-axis); (\textbf{B}) first and second principal components of the hidden states of the full network model. Each dot corresponds to one hidden state and the colors correspond to the input image numeral (0--9); (\textbf{C})~principal components of the composite~model. }
	\label{fig;space}
\end{figure}

We further took the original (continuous-value) hidden states derived from the 60,000 training images and clustered them using $k$-means clustering. We find the optimal number of clusters, identified using the elbow method, to~be 10 or close to 10 (see Figure~\ref{fig;space}A), which of course coincides with the number of image~classes. 

We also performed a principal component analysis (PCA) on the original (not binned) hidden states, and~plot the results for the two first components, color-coding each hidden state by its affiliated class. Again, for~both full and composite networks, we find the hidden states to cluster according to their assigned class (see Figure~\ref{fig;space}B,C). All of these measurements confirm that the effective space of the hidden states is significantly smaller than $2^{20}$ because training causes the hidden states to converge to the attractors needed to classify the images into the ten classes, giving rise to $H_{\rm act}(Y)\approx \log_2(10)$.

\bibstyle{plain}

%\end{adjustwidth}

\begin{thebibliography}{999}

\bibitem[Castelvecchi(2016)]{castelvecchi2016can}
Castelvecchi, D.
\newblock Can we open the black box of AI?
\newblock {\em Nature} {\bf 2016}, {\em 538},~20--223.

\bibitem[Adadi and Berrada(2018)]{adadi2018peeking}
Adadi, A.; Berrada, M.
\newblock Peeking inside the black-box: A survey on explainable artificial
  intelligence (XAI).
\newblock {\em IEEE Access} {\bf 2018}, {\em 6},~52138--52160.

\bibitem[Schreiber(2000)]{schreiber2000measuring}
Schreiber, T.
\newblock Measuring information transfer.
\newblock {\em Phys. Rev. Lett.} {\bf 2000}, {\em 85},~461.

\bibitem[Amblard and Michel(2011)]{amblard2011directed}
Amblard, P.O.; Michel, O.J.
\newblock On directed information theory and Granger causality graphs.
\newblock {\em J. Comput. Neurosci.} {\bf 2011}, {\em
  30},~7--16.

\bibitem[Tehrani-Saleh and Adami(2020)]{tehrani2020can}
Tehrani-Saleh, A.; Adami, C.
\newblock Can transfer entropy infer information flow in neuronal circuits for
  cognitive processing?
\newblock {\em Entropy} {\bf 2020}, {\em 22},~385.

\bibitem[Hintze and Adami(2020)]{hintze2020cryptic}
Hintze, A.; Adami, C.
\newblock Cryptic information transfer in differently-trained recurrent neural
  networks.
\newblock In Proceedings of the 2020 7th International Conference on Soft
  Computing \& Machine Intelligence (ISCMI),  Stockholm, Sweden, 14--15 November  2020; pp.~115--120.

\bibitem[McDonnell \em{et~al.}(2011)McDonnell, Ikeda, and
  Manton]{mcdonnell2011introductory}
McDonnell, M.D.; Ikeda, S.; Manton, J.H.
\newblock An introductory review of information theory in the context of
  computational neuroscience.
\newblock {\em Biol. Cybern.} {\bf 2011}, {\em 105},~55--70.

\bibitem[Dimitrov \em{et~al.}(2011)Dimitrov, Lazar, and
  Victor]{dimitrov2011information}
Dimitrov, A.G.; Lazar, A.A.; Victor, J.D.
\newblock Information theory in neuroscience.
\newblock {\em J. Comput. Neurosci.} {\bf 2011}, {\em
  30},~1--5.

\bibitem[Timme and Lapish(2018)]{timme2018tutorial}
Timme, N.M.; Lapish, C.
\newblock A tutorial for information theory in neuroscience.
\newblock {\em eNeuro} {\bf 2018}, {\em 5}, PMC6131830 .

\bibitem[Bialek \em{et~al.}(2001)Bialek, Nemenman, and
  Tishby]{bialek2001predictability}
Bialek, W.; Nemenman, I.; Tishby, N.
\newblock Predictability, complexity, and learning.
\newblock {\em Neural Comput.} {\bf 2001}, {\em 13},~2409--2463.

\bibitem[Ay \em{et~al.}(2008)Ay, Bertschinger, Der, G{\"u}ttler, and
  Olbrich]{ay2008predictive}
Ay, N.; Bertschinger, N.; Der, R.; G{\"u}ttler, F.; Olbrich, E.
\newblock Predictive information and explorative behavior of autonomous robots.
\newblock {\em  Eur. Phys. J. B} {\bf 2008}, {\em 63},~329--339.

\bibitem[Tononi(2015)]{tononi2015integrated}
Tononi, G.
\newblock Integrated information theory.
\newblock {\em Scholarpedia} {\bf 2015}, {\em 10},~4164.

\bibitem[Fan(2014)]{fan2014information}
Fan, J.
\newblock An information theory account of cognitive control.
\newblock {\em Front. Hum. Neurosci.} {\bf 2014}, {\em 8},~680.

\bibitem[Borst and Theunissen(1999)]{borst1999information}
Borst, A.; Theunissen, F.E.
\newblock Information theory and neural coding.
\newblock {\em Nat. Neurosci.} {\bf 1999}, {\em 2},~947--957.

\bibitem[Marstaller \em{et~al.}(2013)Marstaller, Hintze, and
  Adami]{marstaller2013evolution}
Marstaller, L.; Hintze, A.; Adami, C.
\newblock The evolution of representation in simple cognitive networks.
\newblock {\em Neural Comput.} {\bf 2013}, {\em 25},~2079--2107.

\bibitem[Sporns(2022)]{sporns2022structure}
Sporns, O.
\newblock Structure and function of complex brain networks.
\newblock {\em Dialogues Clin. Neurosci.} {\bf 2022}, {\em
  15},~247--262.

\bibitem[Hagmann \em{et~al.}(2008)Hagmann, Cammoun, Gigandet, Meuli, Honey,
  Wedeen, and Sporns]{hagmann2008mapping}
Hagmann, P.; Cammoun, L.; Gigandet, X.; Meuli, R.; Honey, C.J.; Wedeen, V.J.;
  Sporns, O.
\newblock Mapping the structural core of human cerebral cortex.
\newblock {\em PLoS Biol.} {\bf 2008}, {\em 6},~e159.

\bibitem[Sporns and Betzel(2016)]{SpornsBetzel2016}
Sporns, O.; Betzel, R.F.
\newblock Modular Brain Networks.
\newblock {\em Annu. Rev. Psychol.} {\bf 2016}, {\em 67},~613--40.

\bibitem[Logothetis(2008)]{logothetis2008we}
Logothetis, N.K.
\newblock What we can do and what we cannot do with {fMRI}.
\newblock {\em Nature} {\bf 2008}, {\em 453},~869--878.

\bibitem[He \em{et~al.}(2009)He, Wang, Wang, Chen, Yan, Yang, Tang, Zhu, Gong,
  Zang, et~al.]{he2009uncovering}
He, Y.; Wang, J.; Wang, L.; Chen, Z.J.; Yan, C.; Yang, H.; Tang, H.; Zhu, C.;
  Gong, Q.; Zang, Y.;  et~al.
\newblock Uncovering intrinsic modular organization of spontaneous brain
  activity in humans.
\newblock {\em PLoS ONE} {\bf 2009}, {\em 4},~e5226.

\bibitem[Thatcher(2011)]{thatcher2011neuropsychiatry}
Thatcher, R.W.
\newblock Neuropsychiatry and quantitative EEG in the 21st Century.
\newblock {\em Neuropsychiatry} {\bf 2011}, {\em 1},~495--514.

\bibitem[Shine \em{et~al.}(2021)Shine, Li, Koyejo, Fulcher, and
  Lizier]{shine2021nonlinear}
Shine, J.M.; Li, M.; Koyejo, O.; Fulcher, B.; Lizier, J.T.
\newblock Nonlinear reconfiguration of network edges, topology and information
  content during an artificial learning task.
\newblock {\em Brain Inform.} {\bf 2021}, {\em 8},~1--15.

\bibitem[Hintze \em{et~al.}(2018)Hintze, Kirkpatrick, and
  Adami]{hintze2018structure}
Hintze, A.; Kirkpatrick, D.; Adami, C.
\newblock The structure of evolved representations across different substrates
  for artificial intelligence.
\newblock In Proceedings of the Proceedings Artificial Life 16, Beppu,  Japan, 1--4 February 2018; Ikegami, T.,
  Virgo, N., Witkowski, O., Oka, M., Suzuki, R., Iizuka, H., Eds.; MIT Press:
  Cambridge, MA, USA, 2018.

\bibitem[Kirkpatrick and Hintze(2019)]{kirkpatrick2019role}
Kirkpatrick, D.; Hintze, A.
\newblock The role of ambient noise in the evolution of robust mental
  representations in cognitive systems.
\newblock In Proceedings of the ALIFE 2019: The 2019 Conference on Artificial
  Life, Newcastle-upon-Tyne, UK,  29 July--2 August 2019; MIT Press: Cambridge, MA, USA,   2019; pp. 432--439.

\bibitem[{C G} \em{et~al.}(2018){C G}, Lundrigan, Smale, and
  Hintze]{cg2018effect}
{CG}, N.; Lundrigan, B.; Smale, L.; Hintze, A.
\newblock The effect of periodic changes in the fitness landscape on brain
  structure and function.
\newblock In Proceedings of the ALIFE 2018: The 2018 Conference on Artificial
  Life, Tokyo, Japan, 22--28 July 2018; pp.~469--476.

\bibitem[McCloskey and Cohen(1989)]{mccloskey1989catastrophic}
McCloskey, M.; Cohen, N.J.
\newblock Catastrophic interference in connectionist networks: The sequential
  learning problem. In {\em Psychology of Learning and Motivation}; Elsevier:  Amsterdam, The Netherlands, 
  1989; Volume~24, pp. 109--165.

\bibitem[French(1999)]{french1999catastrophic}
French, R.M.
\newblock Catastrophic forgetting in connectionist networks.
\newblock {\em Trends Cogn. Sci.} {\bf 1999}, {\em 3},~128--135.

\bibitem[Stanley \em{et~al.}(2019)Stanley, Clune, Lehman, and
  Miikkulainen]{stanley2019designing}
Stanley, K.O.; Clune, J.; Lehman, J.; Miikkulainen, R.
\newblock Designing neural networks through neuroevolution.
\newblock {\em Nat. Mach. Intell.} {\bf 2019}, {\em 1},~24--35.

\bibitem[Hintze and Adami(2008)]{hintze2008evolution}
Hintze, A.; Adami, C.
\newblock Evolution of complex modular biological networks.
\newblock {\em PLoS Comput. Biol.} {\bf 2008}, {\em 4},~e23.

\bibitem[Ellefsen \em{et~al.}(2015)Ellefsen, Mouret, and
  Clune]{ellefsen2015neural}
Ellefsen, K.O.; Mouret, J.B.; Clune, J.
\newblock Neural modularity helps organisms evolve to learn new skills without
  forgetting old skills.
\newblock {\em PLoS Comput. Biol.} {\bf 2015}, {\em 11},~e1004128.

\bibitem[Hintze(2021)]{hintze2021role}
Hintze, A.
\newblock The Role Weights Play in Catastrophic Forgetting.
\newblock In Proceedings of the 2021 8th International Conference on Soft
  Computing \& Machine Intelligence (ISCMI), Cairo, Egypt, 26--27 November  2021; pp. 160--166.

\bibitem[Hinton \em{et~al.}(2012)Hinton, Srivastava, Krizhevsky, Sutskever, and
  Salakhutdinov]{hinton2012improving}
Hinton, G.E.; Srivastava, N.; Krizhevsky, A.; Sutskever, I.; Salakhutdinov,
  R.R.
\newblock Improving neural networks by preventing co-adaptation of feature
  detectors.
\newblock {\em arXiv} {\bf 2012}, arXiv:1207.0580.

\bibitem[Parisi \em{et~al.}(2019)Parisi, Kemker, Part, Kanan, and
  Wermter]{parisi2019continual}
Parisi, G.I.; Kemker, R.; Part, J.L.; Kanan, C.; Wermter, S.
\newblock Continual lifelong learning with neural networks: {A} review.
\newblock {\em Neural Netw.} {\bf 2019}, {\em 113},~54--71.

\bibitem[Kirkpatrick \em{et~al.}(2017)Kirkpatrick, Pascanu, Rabinowitz, Veness,
  Desjardins, Rusu, Milan, Quan, Ramalho, Grabska-Barwinska,
  et~al.]{kirkpatrick2017overcoming}
Kirkpatrick, J.; Pascanu, R.; Rabinowitz, N.; Veness, J.; Desjardins, G.; Rusu,
  A.A.; Milan, K.; Quan, J.; Ramalho, T.; Grabska-Barwinska, A.;  et~al.
\newblock Overcoming catastrophic forgetting in neural networks.
\newblock {\em Proc. Natl. Acad. Sci. USA} {\bf 2017}, {\em
  114},~3521--3526.

\bibitem[Ribeiro \em{et~al.}(2016)Ribeiro, Singh, and
  Guestrin]{ribeiro2016should}
Ribeiro, M.T.; Singh, S.; Guestrin, C.
\newblock {``Why should I trust you?'' Explaining} the predictions of any
  classifier.
\newblock In Proceedings of the 22nd ACM SIGKDD
  International Conference on Knowledge Discovery and Data Mining,  San Francisco, CA, USA, 13--17 August 2016; pp.~1135--1144.

\bibitem[Golden \em{et~al.}(2022)Golden, Delanois, Sanda, and
  Bazhenov]{Goldenetal2022}
Golden, R.; Delanois, J.E.; Sanda, P.; Bazhenov, M.
\newblock Sleep prevents catastrophic forgetting in spiking neural networks by
  forming a joint synaptic weight representation.
\newblock {\em PLoS Comput. Biol.} {\bf 2022}, {\em 18},~e1010628.

\bibitem[Kemker \em{et~al.}(2018)Kemker, McClure, Abitino, Hayes, and
  Kanan]{kemker2018measuring}
Kemker, R.; McClure, M.; Abitino, A.; Hayes, T.; Kanan, C.
\newblock Measuring catastrophic forgetting in neural networks.
\newblock In Proceedings of the  32nd AAAI Conference on
  Artificial Intelligence,  New Orleans, LA, USA, 2--7 February 2018; pp. 3390--3398.

\bibitem[Bohm \em{et~al.}(2022)Bohm, Kirkpatrick, Cao, and
  Adami]{bohm2022information}
Bohm, C.; Kirkpatrick, D.; Cao, V.; Adami, C.
\newblock Information fragmentation, encryption and information flow in complex
  biological networks.
\newblock {\em Entropy} {\bf 2022}, {\em 24},~735.

\bibitem[Sella(2022)]{sella2022tracing}
Sella, M.
\newblock Tracing Computations in Deep Neural Networks.
\newblock Master's Thesis, School of Information and Engineering, Dalarna
  University, Falun, Sweden,  2022.

\bibitem[Paszke \em{et~al.}(2019)Paszke, Gross, Massa, Lerer, Bradbury, Chanan,
  Killeen, Lin, Gimelshein, Antiga, Desmaison, Kopf, Yang, DeVito, Raison,
  Tejani, Chilamkurthy, Steiner, Fang, Bai, and Chintala]{Paszkeetal2019}
Paszke, A.; Gross, S.; Massa, F.; Lerer, A.; Bradbury, J.; Chanan, G.; Killeen,
  T.; Lin, Z.; Gimelshein, N.; Antiga, L.;  et~al.
\newblock PyTorch: {An} Imperative Style, High-Performance Deep Learning
  Library.
\newblock In Proceedings of the Advances in Neural Information Processing
  Systems, Vancouver, BC, Canada, 8--14 December 2019; Wallach, H., Larochelle, H., Beygelzimer, A., d'Alch\'{e}-Buc, F., Fox, E., Garnett, R., Eds.; Curran Associates, Inc.:  Red Hook, NY, USA, 2019,
  Volume~32.

\bibitem[LeCun \em{et~al.}(1998)LeCun, Bottou, Bengio, and
  Haffner]{lecun1998gradient}
LeCun, Y.; Bottou, L.; Bengio, Y.; Haffner, P.
\newblock Gradient-based learning applied to document recognition.
\newblock {\em Proc. IEEE} {\bf 1998}, {\em 86},~2278--2324.

\bibitem[Kingma and Ba(2015)]{kingma2014adam}
Kingma, D.P.; Ba, J.
\newblock Adam: {A} method for stochastic optimization.
\newblock In Proceedings of the 3rd International Conference for Learning
  Representations, San Diego, CA, USA, 7--9 May 2015; Bengio, Y., LeCun, Y., Eds.;   2015.

\bibitem[Shannon(1948)]{shannon1948mathematical}
Shannon, C.E.
\newblock A mathematical theory of communication.
\newblock {\em  Bell Syst. Tech. J.} {\bf 1948}, {\em
  27},~379--423.

\bibitem[Paninski(2003)]{Paninski2003}
Paninski, L.
\newblock Estimation of entropy and mutual information.
\newblock {\em Neural Comput.} {\bf 2003}, {\em 15},~1191--1253.

\bibitem[Bohm \em{et~al.}(2022)Bohm, Kirkpatrick, and
  Hintze]{bohm2022understanding}
Bohm, C.; Kirkpatrick, D.; Hintze, A.
\newblock Understanding memories of the past in the context of different
  complex neural network architectures.
\newblock {\em Neural Comput.} {\bf 2022}, {\em 34},~754--780.

\bibitem[Chapman \em{et~al.}(2013)Chapman, Knoester, Hintze, and
  Adami]{chapman2013evolution}
Chapman, S.; Knoester, D.; Hintze, A.; Adami, C.
\newblock Evolution of an artificial visual cortex for image recognition.
\newblock In Proceedings of the ECAL 2013: The Twelfth European Conference on
  Artificial Life,   Taormina, Italy, 2--6 September 2013; pp. 1067--1074.

\bibitem[Basharin(1959)]{Basharin1959}
Basharin, G.P.
\newblock On a statistical estimate for the entropy of a sequence of
  independent random variables.
\newblock {\em Theory Probab. Applic.} {\bf 1959}, {\em 4},~333--337.
\end{thebibliography}
\end{document}